\def\year{2019}\relax
\documentclass[letterpaper]{article} 
\usepackage{aaai19}  
\usepackage{times}  
\usepackage{helvet}  
\usepackage{courier}  
\usepackage{url}  
\usepackage{graphicx}  

\usepackage{amsmath,amssymb,amsthm}
\usepackage{mathabx}
\usepackage{multirow}
\usepackage{url}
\usepackage{hhline}
\usepackage{tikz,pgfplots}
\usepackage{xcolor}
\usetikzlibrary{arrows,intersections}
\usepackage{multirow}
\usepackage{pgfplots}
\usepackage{algorithm}
\usepackage[noend]{algpseudocode}

\usepackage{latexsym}

\usepackage{color}
\usepackage{multirow}
\usepackage{amsfonts}
\usepackage{float}
\usepackage{graphicx}
\usepackage{graphics}
\usepackage[export]{adjustbox}
\usepackage{adjustbox}
\usepackage{booktabs}
\usepackage{array}

\usepackage{pdfpages}
\usepackage{arydshln}

\usepackage{subcaption}
\usepackage{multirow}
\include{plots}
\definecolor{bblue}{HTML}{4F81BD}
\definecolor{rred}{HTML}{C0504D}
\definecolor{ggreen}{HTML}{9BBB59}
\definecolor{ppurple}{HTML}{9F4C7C}
\definecolor{Dark scarlet}{HTML}{560319}
\definecolor{Forest green}{HTML}{1E4D2B}

\newcounter{notecounter}
\newcommand{\enotesoff}{\long\gdef\enote##1##2{}}

\enotesoff

\def\dnrm#1{\mbox{$_{\hbox{\scriptsize #1}}$}}

\def\citet#1{\citeauthor{#1} \shortcite{#1}}

\newcommand*{\affaddr}[1]{#1} 
\newcommand*{\affmark}[1][*]{\textsuperscript{#1}}
\newcommand*{\email}[1]{\texttt{#1}}

\begin{document}
%

  \pdfinfo{
/Title (Neural Relation Extraction Within and Across Sentence Boundaries)
/Author (Pankaj Gupta, Subburam Rajaram, Hinrich Sch\"{u}tze, Thomas Runkler)}
\setcounter{secnumdepth}{0}  
%
\title{Neural Relation Extraction Within and Across Sentence Boundaries}

\author{Pankaj Gupta\affmark[1,2], Subburam Rajaram\affmark[1], Bernt Andrassy\affmark[1], Hinrich Sch\"{u}tze\affmark[2], Thomas Runkler\affmark[1]\\ 
 \affaddr{\affmark[1]Corporate Technology, Machine-Intelligence (MIC-DE), Siemens AG  Munich, Germany}\\
  \affaddr{\affmark[2]CIS, University of Munich (LMU) Munich, Germany} \\
  \email{\{pankaj.gupta, subburam.rajaram\}@siemens.com} $|$ 
  \email{ pankaj.gupta@campus.lmu.de}
}

\maketitle

\begin{abstract}

Past work in relation extraction mostly focuses on binary
relation between entity pairs \emph{within single sentence}.
Recently, the NLP community has gained interest in relation
extraction in entity pairs \emph{spanning multiple
  sentences}.  In this paper, we propose
a novel architecture for this task:
inter-sentential
dependency-based neural networks (iDepNN).
iDepNN models the
shortest and augmented dependency paths via recurrent and
recursive neural networks to extract relationships within
(intra-) and across (inter-) sentence boundaries.  Compared
to SVM and neural network baselines, iDepNN is more robust to false positives
in relationships spanning sentences.  We evaluate our models
on four datasets from newswire (MUC6) and medical (BioNLP
shared task) domains that achieve state-of-the-art
performance and show a better balance in precision and
recall for inter-sentential relationships. 
We perform better than 11 teams participating in 
the BioNLP shared task 2016 
and achieve a gain of $5.2$\% ($0.587$ vs $0.558$) in $F_1$ over the winning team. 
 We also release the
cross-sentence annotations for MUC6.
\end{abstract}

\section{Introduction}
\label{introduction}



The task of relation extraction (RE) aims to identify semantic relationship between a pair of nominals or entities \textit{e1} and \textit{e2} in a given sentence S. 
Due to a rapid growth in information, it plays a vital role in knowledge extraction from unstructured texts and serves as an intermediate 
step in a variety of NLP applications in  
newswire, web and high-valued biomedicine \cite{Bahcall:82} domains. 
Consequently, there has been increasing interest in relation extraction, particularly in augmenting existing knowledge bases.  

\begin{figure*}[t]
{
  \centering
   \includegraphics[clip,max width=0.96\textwidth]{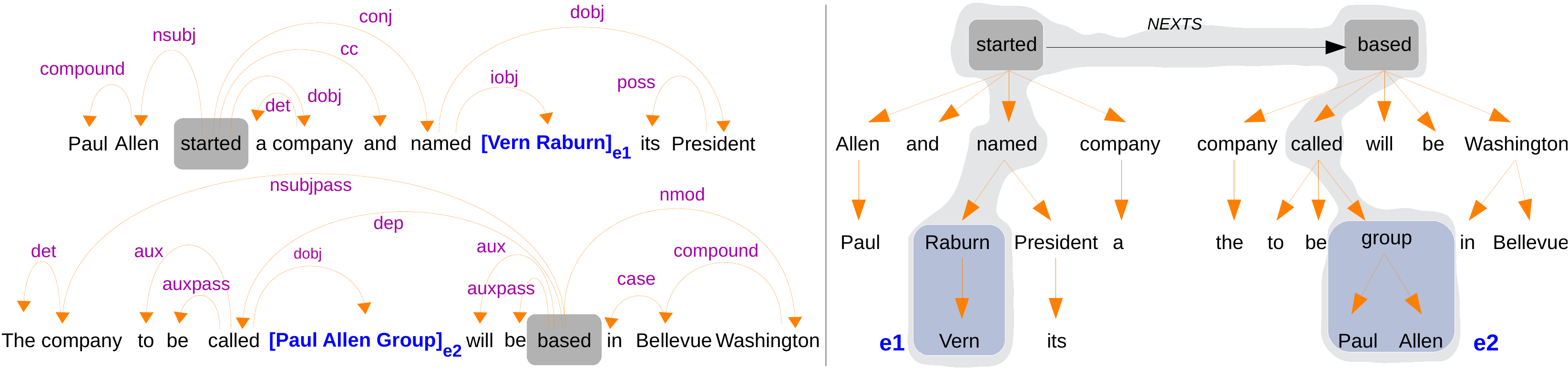}
    \caption{Left: Sentences and their dependency graphs. Right: Inter-sentential Shortest Dependency Path (iSDP) across sentence boundary. Connection between the roots of adjacent sentences by {\it NEXTS}.}
    \label{fig:sdp_illustration.}
}
\end{figure*}

Progress in relation extraction is exciting; however most  prior work \cite{zhang2006composite,kambhatla2004combining,vu2016combining,gupta2016table}  is limited to 
single sentences, i.e., {\it intra-sentential} relationships, and ignores relations in entity pairs spanning sentence boundaries, i.e., {\it inter-sentential}.  
Thus, there is a  need to move beyond single sentences and devise methods to extract relationships spanning sentences.  
For instance, consider the 
sentences:

{\small 
Paul Allen has started a company and named 
$\mbox{[{\it Vern}}$ $\mbox{\it Raburn}]_{\it e1}$ its President.
The company, to be called $\mbox{[{\it Paul Allen
Group}]}_{\it e2}$ will be based in Bellevue, Washington.} 

The two sentences together convey the fact that the entity
\emph{e1} is associated with \emph{e2}, which cannot be inferred from
either sentence alone.  The missed relations impact the
system performance, leading to poor recall.
But precision is equally important; e.g., 
in high-valued
biomedicine domain, 
significant inter-sentential
relationships must be extracted, especially in  medicine that aims
toward accurate diagnostic testing and precise treatment,
and extraction errors can have severe negative consequences. In
this work, we present a neural network (NN) based approach to
precisely extract relationships within and across sentence
boundaries, and show a better balance in precision and
recall with an improved $F_1$.


Previous work on cross-sentence relation extraction used
coreferences to access entities that occur in a different
sentence \cite{gerber2010beyond,yoshikawa2011coreference}
without modeling inter-sentential relational patterns. 
\citeauthor{swampillai2011extracting} \shortcite{swampillai2011extracting} described a SVM-based
approach to both intra- and inter-sentential relations.
Recently, \citeauthor{quirk2016distant} \shortcite{quirk2016distant} 
applied distant 
supervision to cross-sentence relation extraction of 
entities using binary logistic regression (non-neural
network based) classifier and \citeauthor{peng2017cross} \shortcite{peng2017cross} 
applied sophisticated graph long short-term memory networks
to cross-sentence n-ary relation extraction.  However, it
still remains challenging due to the need for coreference
resolution, noisy text between the entity pairs spanning
multiple sentences and lack of labeled corpora.


\citeauthor{bunescu2005shortest} \shortcite{bunescu2005shortest}, \citeauthor{nguyen2007relation} \shortcite{nguyen2007relation} and \citeauthor{mintz:82} \shortcite{mintz:82} 
have shown that the shortest dependency path (SDP)
between two entities in a dependency graph and the dependency subtrees are the most useful dependency features   
in relation classification. 
Further, \citeauthor{liu2015dependency} \shortcite{liu2015dependency} 
developed these ideas using Recursive Neural Networks (\emph{RecNNs}, \citet{socher2014grounded}) and combined the two components in a precise structure called 
Augmented Dependency Path (ADP), where each word on a SDP is attached to a dependency subtree; however, limited to single sentences. 
In this paper, we aspire from these methods to extend shortest dependency path across sentence boundary and effectively combine it with dependency subtrees in NNs that 
can capture semantic representation of the structure and boost relation extraction spanning sentences. 

The {\it contributions} are: 
{\bf (1)} Introduce a novel dependency-based neural architecture, named as inter-sentential Dependency-based Neural Network (iDepNN) 
to extract relations within and across sentence boundaries by modeling shortest and augmented dependency paths 
in a combined structure of bidirectional RNNs (biRNNs) and RecNNs.
{\bf (2)}  Evaluate different linguistic features on four datasets from newswire and  medical domains, and report an improved performance in relations spanning sentence boundary.  
We show amplified precision due to robustness towards false positives, and a better balance in precision and recall. 
We perform better than 11 teams participating in 
 in the BioNLP shared task 2016  
and achieve a gain of $5.2$\% ($0.587$ vs $0.558$) in $F_1$ over the winning team.  
{\bf (3)} Release relation annotations for the MUC6 dataset
for intra- and inter-sentential relationships. 
{\it Code}, {\it data} and {\it supplementary} are available at  
{\small \url{https://github.com/pgcool/Cross-sentence-Relation-Extraction-iDepNN}}. 

\section{Methodology}
\label{methodology}

\subsection{Inter-sentential Dependency-Based Neural Networks (iDepNN)}
Dependency-based neural networks
(DepNN) \cite{bunescu2005shortest,liu2015dependency} have
been investigated for relation extraction between entity
pairs limited to single sentences, using the dependency
information to explore the semantic connection between two
entities.  In this work, we
introduce
\emph{iDepNN}, the
inter-sentential Dependency-based Neural
Network, an NN that models
relationships between entity pairs spanning sentences, i.e.,
inter-sentential within a document. 
We refer to the iDepNN that only
models the shortest dependency path (SDP) spanning sentence
boundary  as \emph{iDepNN-SDP} and to the iDepNN
that models 
augmented dependency paths (ADPs)  as \emph{iDepNN-ADP}; see below.
biRNNs (bidirectional
RNNs, \citet{schuster1997bidirectional}) and
RecNNs (recursive NNs, \citet{socher2012semantic}) are the
backbone of iDepNN.

\begin{figure*}[t]
{
  \centering
   \includegraphics[scale=.65]{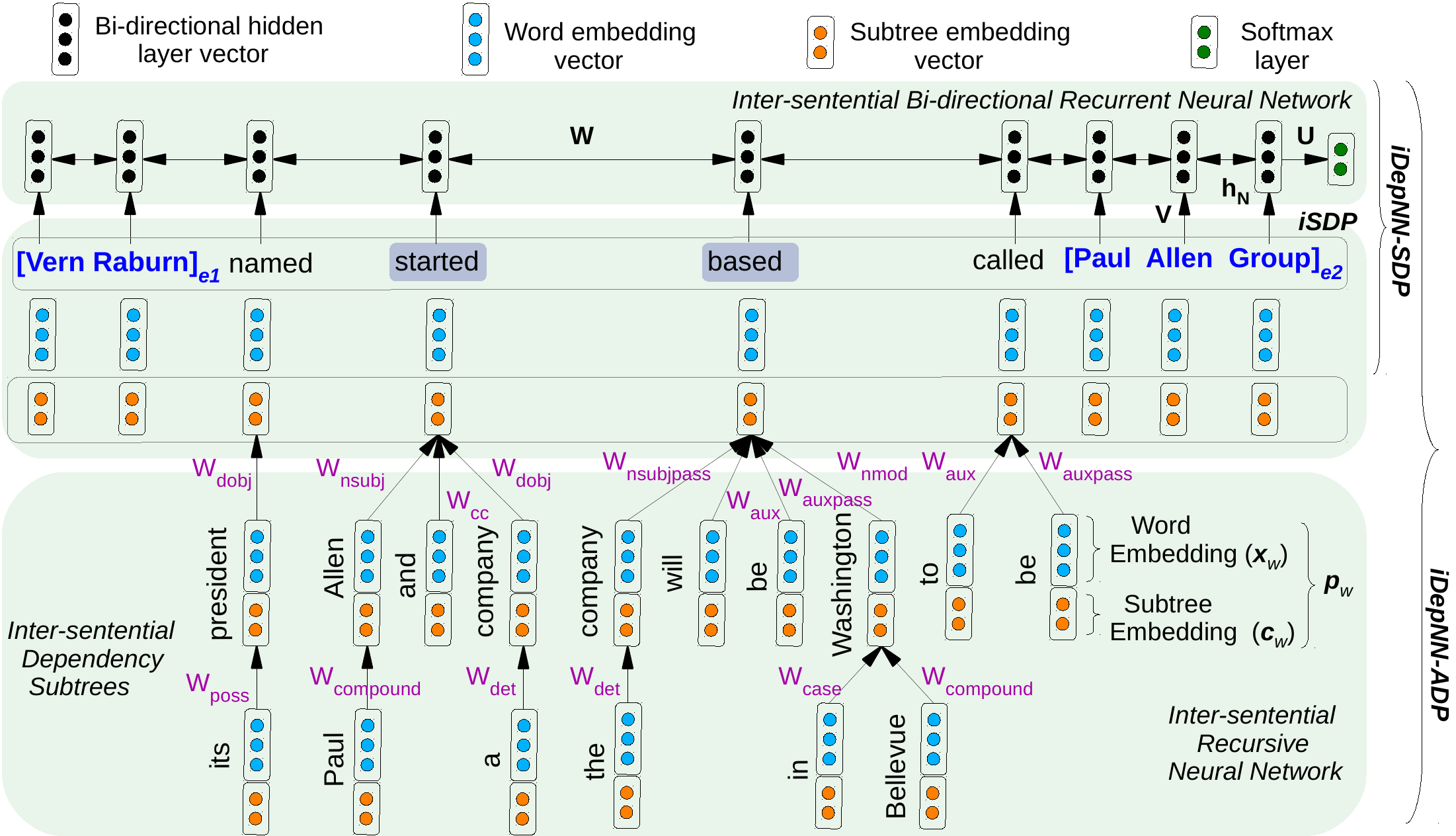}
    \caption{Inter-sentential Dependency-based Neural Network variants: iDepNN-SDP and iDepNN-ADP}
    \label{fig:isdpnetwork}
}
\end{figure*}

{\bf Modeling Inter-sentential Shortest Dependency Path (iDepNN-SDP)}:  
We compute the inter-sentential Shortest Dependency Path (iSDP) between entities spanning sentence boundaries for a relation. 
To do so, we obtain the dependency parse tree for each sentence using the StanfordCoreNLP dependency parser \cite{manning2014stanford}. 
We then use NetworkX \cite{hagberg2008exploring} to represent each token as a node and the dependency relation as a link between the nodes. 
In the case of multiple sentences, the root node of the parse tree of a sentence is connected to the root of the subsequent tree, leading to the shortest path from one entity  to another across sentences.

Figure \ref{fig:sdp_illustration.} (Left) shows dependency
graphs for the example sentences where the two
entities \emph{e1} and \emph{e2} appear in nearby sentences
and exhibit a relationship.
Figure \ref{fig:sdp_illustration.} (Right) illustrates that
the dependency trees of the two adjacent sentences and their
roots are connected by {\it NEXTS} to form an \emph{iSDP},
an inter-Sentential Dependency Path, (highlighted in gray)
between the two entities.  The shortest path spanning
sentence boundary is seen as a sequence of words between two
entities. Figure \ref{fig:isdpnetwork} shows how a
biRNN
\cite{schuster1997bidirectional,vu2016bi} uses
iSDP to detect relation between 
\emph{e1}
and \emph{e2}, positioned one sentence apart.

{\bf Modeling Inter-sentential Dependency Subtrees}: 
To effectively represent words on the shortest dependency path within and across sentence boundary, 
we model dependency subtrees assuming that
each word $w$ can be seen as the word itself and its
children on the dependency subtree. The notion of
representing words using subtree vectors within the
dependency neural network  (DepNN) is similar to
\cite{liu2015dependency}; 
however, 
our proposed structures are based on iSDPs and ADPs that span sentences. 

\enote{hs}{suggestion: put braces next to iSDP and ADP in Figur 2}

To represent each word $w$ on the subtree, its word
embedding vector ${\bf x}_w \in {\cal R}^{d}$ and subtree
representation ${\bf c}_w \in {\cal R}^{d'}$ are
concatenated to form its final representation ${\bf p}_w \in
{\cal R}^{d+d'}$.  We use 
200-dimensional pretrained
GloVe embeddings \cite{pennington2014glove}.
The subtree representation of a word is
computed through recursive transformations of the
representations of its children words.  A RecNN
is used to construct subtree embedding ${\bf c}_w$,
traversing bottom-up from its leaf words to the root for
entities spanning sentence boundaries, as shown in
Figure \ref{fig:isdpnetwork}.  For a word which is a leaf
node, i.e., it does not have a subtree, we set its subtree
representation as ${\bf c}\dnrm{LEAF}$.
Figure
~\ref{fig:isdpnetwork} illustrates how 
subtree-based word
representations are constructed via iSDP.



Each word is associated with a dependency relation $r$,
e.g.,  $r$ = \emph{dobj},
during the bottom-up construction of the subtree. 
For each  $r$, a transformation matrix ${\bf W}_r \in {\cal R}^{d' \times (d+d')}$ 
is  learned. The subtree embedding is computed as:
\small \[
{\bf c}_w = f( \sum_{q \in Children(w)}  {\bf W}_{R_{(w,q)}} \cdot {\bf p}_q + {\bf b})  \ \mbox{and} \  {\bf p}_q  = [{\bf x}_q, {\bf c}_q] 
\]\normalsize
where $R_{(w,q)}$ is the dependency relation between
word $w$ and its child word $q$ and ${\bf b} \in {\cal R}^{d'}$ is a bias.  
This process continues recursively up to the root word such
as the word ``named'' on the iSDP in the figure.
{\bf Modeling Inter-sentential Augmented Dependency Path (iDepNN-ADP)}: 
Following \citeauthor{liu2015dependency} \shortcite{liu2015dependency}, 
we combine the two components: iSDP and {\it dependency subtrees} spanning sentence boundaries 
to form a combined structure which we name as  inter-sentential Augmented Dependency Path ({\it iDepNN-ADP}). 
As shown in Figure \ref{fig:isdpnetwork}, each word on iSDP is attached to its subtree representation ${\bf c}_w$.  
An attached subtree enriches each word on the iSDP with additional information about how this word functions in specific sentence to form a more precise structure 
for classifying relationships within and across sentences.

To capture the semantic representation of {\it iDepNN-ADP}, we first adopt a RecNN to model the dependency subtrees for each word on the iSDP. 
Then, we design a biRNN to obtain salient semantic features  on the iSDP. The overall structure of {\it iDepNN-ADP} (Figure \ref{fig:isdpnetwork}) is built upon the combination of recursive and recurrent NNs spanning sentences. 

{\bf Learning}: We develop a biRNN over the two structures:
iDepNN-SDP and iDepNN-ADP, and pass the last hidden vector
$\bf h_N$ (in the iSDP word sequence,
Figure \ref{fig:isdpnetwork}) to a softmax layer whose
output is the probability distribution $\bf y$ over relation
labels $R$, as ${\bf y} = \mbox{softmax}({\bf U} \cdot {\bf h}_N +
{\bf b}_y)$ where ${\bf U} \in {\cal R}^{R \times H}$ is the weight
matrix connecting hidden vector of dimension $H$ to output
of dimension $R$ and ${\bf b}_y \in {\cal R}^R$ is the
bias. ${\bf h}_N$ is the last hidden vector of the biRNN.


To compute semantic representation ${\bf h}_w$ for each word $w$ on the iSDP, we adopt the {\it Connectionist biRNN} \cite{vu2016combining} that
combines the forward and backward pass by adding their hidden layers (${\bf h}_{{f}_{t}}$  and ${\bf h}_{{b}_{t}}$) at each time step $t$ and also adds a weighted connection to the
previous combined hidden layer $ {\bf h}_{t-1}$ to include all intermediate hidden layers into the final decision.
\begin{eqnarray*}
{\bf h}_{{f}_{t}} = f({\bf V} \cdot {\bf i}_{t} + {\bf W} \cdot  {\bf h}_{{f}_{t-1}})\\
{\bf h}_{{b}_{t}} = f({\bf V} \cdot {\bf i}_{N-t+1} + {\bf W} \cdot {\bf h}_{{b}_{t+1}})\\
{\bf h}_{t} = f({\bf h}_{{f}_{t}} + {\bf h}_{{b}_{t}}+ {\bf W} \cdot {\bf h}_{t-1})
\end{eqnarray*}
where ${\bf V} \in {\cal R}^{H \times |{\bf i}|}$, $N$ is the total number of words on iSDP and ${\bf i}_t$  the input vector at  $t$, defined by:
\[ \mbox{iDepNN-SDP}: {\bf i}_t \!=\! [{\bf x}_t, {\bf
L}_t] \ \ \ \ \ \ \mbox{iDepNN-ADP}: {\bf i}_t \!=\! [{\bf p}_t, {\bf L}_t] \]
where ${\bf L}_t$ are lexical level features 
(e.g., part-of-speech  tag, position indicators, entity types) for each word at $t$.
Observe, in order to minimize the number of parameters, we share the same weight matrix ${\bf W}$ in three parts: forward pass, backward pass and combination of  both.  
The optimization objective is to minimize
the cross-entropy error between the ground-truth
label and softmax output.
The parameters
are learned using backpropagation \cite{werbos1990backpropagation}.

\enote{hs}{you say here that you use backpropagation, but i
don't think  you say it elsewhere. do you not use
backpropagation elsewhere?}


{\bf Key Features}: 
The features focus on characteristics of the full sentence, dependency path or individual entities. 
The various features used in our experiments are: 
{(1)} {\it Position-Indicator} (PI): A one-hot vector for SVM which indicates the position of the entity in the vocabulary. Four additional words ($<$$e_1$$>$, $<$$/e_1$$>$, $<$$e_2$$>$, $<$$/e_2$$>$)  
to mark start and end  of entity mentions \emph{e1} and \emph{e2},  used in NNs. See details about PI in \citeauthor{Gupta:Thesis:2015} \shortcite{Gupta:Thesis:2015}. 
{(2)} {\it Entity Types} (ET): A one-hot vector to represent the entity type in SVM and embedding vectors in NNs.  
{(3)} {\it Part-of-speech} (POS):  A bag-of-words (BoW) in SVM and embedding vector  for each POS type in NNs.  
{(4)} {\it Dependency}:  In SVM, the specific edge types in the dependency path are captured with a BoW vector, similar to \citeauthor{grouin2016identification}  \shortcite{grouin2016identification}. 
In NNs, it refers to {\it iDepNN-ADP}. 
{(5)} [{\it inter-sentential}-]{\it Shortest-Dependency-Path} ([i-]SDP): Sequence of Words on the [i-]SDP.

\begin{table}[t]
\centering
\def\arraystretch{1.1}
\resizebox{0.48\textwidth}{!}{%
\begin{tabular}{lcc|ccc}
{\it Relation}     & {\it Intra} & {\it Inter} & {\it Relation}     & {\it Intra} & {\it Inter} \\
\hline
\multicolumn{3}{c|}{BioNLP ST 2011 ({\it Medical})}      & \multicolumn{3}{c}{BioNLP ST 2013 ({\it Medical})}      \\
\cline{2-2}\cline{5-5}
{\textit PartOf}       & 99             & 103         & {\textit PartOf }      & 104            & 83            \\
{\textit Localization} & 261            & 732       & {\textit Localization} & 246            & 677        \\
\hline
Total        & 360            & 835 ({\bf 70\%})   & Total & 350            & 760 ({\bf 69\%})  \\  \hline
\multicolumn{3}{c|}{BioNLP ST 2016 ({\it Medical})}    &     \multicolumn{3}{c}{MUC6 ({\it News})}           \\
\cline{2-2}\cline{5-5}
{\textit Lives\_In}    & 363            & 135          & {\textit Per-Org }      & 245            & 112             \\
& & &{\textit Per-Post} & 407            & 66          \\
& & & {\textit Org-Post} & 268            & 113            \\
\hline
Total        & 363            & 135 ({\bf 27\%})   & Total & 920            & 291 ({\bf 24\%})\\ \hline
\end{tabular}}
\caption{Count of intra- and inter-sentential relationships in datasets (train+dev) from two domains}
\label{Intra-sentence and Inter-sentence relationships statistics in different datasets in medical and News article domains.}
\end{table}

\section{Evaluation and Analysis}\label{evaluation_and_analysis}

\textbf{Dataset.}
We evaluate our proposed methods on four datasets from medical and news domain.
Table  \ref{Intra-sentence and Inter-sentence relationships statistics in different datasets in medical and News article domains.} shows  counts of intra- and inter-sentential relationships.
The three medical domain datasets are taken from the BioNLP shared task (ST) of relation/event extraction \cite{bossy2011bionlp,nedellec2013overview,deleger2016overview}.  
We compare our proposed techniques with the systems published at these venues.  
The Bacteria Biotope task \cite{bossy2011bionlp} of the BioNLP ST 2011 focuses on extraction of habitats of bacteria, which is extended by 
the BioNLP ST 2013 \cite{nedellec2013overview}, while  the BioNLP ST 2016 focuses on extraction of \textit{Lives\_in} events. 
We have standard train/dev/test splits for the BioNLP ST 2016 dataset, while we perform 3-fold crossvalidation\footnote{the official evaluation is not accessible any more and therefore, the annotations for their test sets are not available} on BioNLP ST 2011 and 2013. 
For BioNLP ST 2016, we generate negative examples by randomly sampling co-occurring entities without known interactions. Then 
we sample the same number as positives to obtain a balanced dataset  during training and validation for different sentence range. 
See supplementary for further details.

The MUC6 \cite{grishman1996message} dataset contains information about management succession events from newswire. 
The task organizers provided a training corpus and a set of templates that contain the management succession events,    
the names of people who are starting or leaving management posts, the names of their respective posts and organizations and whether the named person is currently in the job. 
\textbf{Entity Tagging}: We tag entities {\it Person} (\textit{Per}), {\it Organization} (\textit{Org}) using Stanford NER tagger \cite{finkel2005incorporating}. 
The entity type {\it Position} (\textit{Post}) is annotated based on the templates. 
\textbf{Relation Tagging}: We have three types of relations: \textit{Per-Org}, \textit{Per-Post} and \textit{Post-Org}. 
We follow \citeauthor{swampillai2010inter} \shortcite{swampillai2010inter} and
annotate binary relations (within and across sentence boundaries) 
using management succession events between two entity pairs. 
We randomly split the collection
 60/20/20 into train/dev/test.


\begin{table}[t]
\centering
\def\arraystretch{1.2}
\resizebox{0.45\textwidth}{!}{%
\begin{tabular}{l|ccc|ccc}
\hline
\multicolumn{7}{c}{{\bf Dataset:} \underline{BioNLP ST 2016}}\\
\multirow{2}{*}{\bf Features}   & \multicolumn{3}{c|}{\bf SVM} & \multicolumn{3}{c}{\bf iDepNN}\\ \cline{2-7}
                      & P      & R     & $F_1$     & P       & R      & $F_1$      \\ \hline
iSDP                   &  .217      &    .816   & .344      &   .352      &     .574   &    .436     \\
+ PI + ET                  &   .218     &   .819    &    .344    &   .340      &  .593      &  .432       \\
+ POS             &   .269     &   .749    &    .396    &   .348      &  .568      &  .431       \\
+ Dependency          &   .284     &   .746    &    .411    &  {.402}      &   {.509}      &  {\bf .449}\\ 
\hline                                                           
\multicolumn{7}{c}{{\bf Dataset:} \underline{MUC6}}\\ 
\multirow{2}{*}{\bf Features}   & \multicolumn{3}{c|}{\bf SVM} & \multicolumn{3}{c}{\bf iDepNN} \\ \cline{2-7}
                      & P      & R     & $F_1$     & P       & R      & $F_1$      \\ \hline
iSDP                  & .689	& .630	& .627           & .916	& .912	& .913     \\               
+ PI                  & .799	& .741	& .725            & .912	& .909	& .909        \\
+ POS         & .794	& .765	& .761               & .928	& .926	& .926\\
+ Dependency             & .808	& .768	& .764              & .937	& .934	& {\bf .935}       \\ \hline     
\end{tabular}}
\caption{SVM vs iDepNN: Features in inter-sentential ($k$$\le$$1$) training and inter-sentential ($k$ $\le$ $1$) evaluation. iSDP+Dependency refers to iDepNN-ADP structure.}
\label{SVMiDepNNFeatureAnalysis}
\end{table}

\begin{table*}[t]
\centering
\def\arraystretch{1.2}
\resizebox{.98\textwidth}{!}{%
\begin{tabular}{c|c||cccc|cccc|cccc|cccc}
\multirow{1}{*}{\bf train}   & \multirow{3}{*}{\bf Model}  &  \multicolumn{16}{c}{\bf Evaluation for different values of sentence range $k$}\\ \cline{3-18}
 \multirow{1}{*}{\bf param}     &   & \multicolumn{4}{c|}{\bf $k=0$} & \multicolumn{4}{c|}{\bf $k \le 1$} & \multicolumn{4}{c|}{\bf $ k \le 2$} & \multicolumn{4}{c}{$k \le 3$}\\ 
        & & ${\bf pr}$ & $P$      & $R$     & $F_1$     & ${\bf pr}$    & $P$       &  $R$      & $F_1$     & ${\bf pr}$     & $P$       & $R$      & $F_1$    & ${\bf pr}$    & $P$      & $R$      & $F_1$ \\ \hline
\multirow{5}{*}{$k=0$}   &  SVM       &  363   &    .474     &   .512    &    .492    &  821   &   .249      &      .606  &  .354  &  1212   &    .199     &  .678      &   .296   &  1517   &  .153  &  .684  &  .250 \\ 
  &  graphLSTM       &  473   &    .472     &   .668    &    .554    &  993   &   .213      &      .632  &  .319  &  1345   &    .166     
&  .660      &   .266   &  2191   &  .121  &  .814  &  .218  \\
\cdashline{2-18}
&  i-biLSTM     &  480  &     {.475}     &   .674    &    .556    &  998   &   .220      &  .652      &  .328   &  1376 &   .165      &  .668      &  .265     &  1637 &  .132  &  .640  &  .219  \\
&  i-biRNN     &  286  &     {.517}     &   .437    &    .474    &  425   &   .301      &  .378      &  .335   &  540 &   .249      &  .398      &  .307     &  570 &  .239  &  .401  &  .299  \\  
& iDepNN-SDP     &  297   &    .519     &   .457    &    .486   &  553   &   .313      &  .510      &  .388  &  729   &   .240      &  .518      &  .328   &  832   &  .209  &  .516  & .298 \\ 
& iDepNN-ADP      &  266   &     \underline{.526}     &   .414    &    .467    &  476   &    \underline{.311}      &  .438      &   .364  &  607   &   \underline{.251}      &  .447      &   .320    &  669   &   {.226}  &  .447  &  .300\\ 
\hline
\multirow{5}{*}{$k \le 1$}   &  SVM      &  471   &    .464     &   .645    &    .540   &  888   &   .284      &      .746  &  .411 &  1109   &    .238     &  .779      &   .365   &  1196   &  .221  &  .779  &  .344  \\ 
  &  graphLSTM       &  406   &    .502     &   .607    &    .548    &  974   &   .226      &      .657  &  .336  &  1503   &    .165     
&  .732      &   .268   &  2177   &  .126  &  .813  &  .218  \\
\cdashline{2-18}
&  i-biLSTM     &  417  &     {.505}     &   .628    &    .556    &  1101   &   .224      &  .730      &  .343   &  1690 &   .162      &  .818      &  .273     &  1969 &  .132  &  .772  &  .226  \\
&  i-biRNN     &  376 &     .489     &   .544    &    .515    &  405  &   .393      &  .469      &  .427  &  406    &   .391      &  .469      &  .426    &  433    &  .369  &  .472  &  .414  \\ 
& iDepNN-SDP    &  303  &    .561     &   .503    &    .531    &  525   &   .358      &  .555      &  .435  &  660   &   .292      &  .569      &  .387    &  724   &  .265  &  .568  &  .362  \\ 
& iDepNN-ADP  &  292   &     \underline{\bf .570}     &   .491    &    .527    &  428  &   \underline{\bf .402}      &  .509      &  {\bf .449}  &  497   &   \underline{\bf .356}      &  .522      &  {\bf .423}    &  517  &  {\bf .341}  &  .521  &  {\bf .412} \\ 
\hline
\multirow{5}{*}{$k \le 2$}   &  SVM       &  495  &    .461     &   .675    &    .547    &  1016   &   .259      &      .780  &  .389  &  1296  &    .218     &  .834      &   .345  &  1418   &  .199  &  .834  &  .321 \\ 
  &  graphLSTM       &  442   &    .485     &   .637    &    .551    &  1016   &   .232      &      .702  &  .347  &  1334   &    .182     
&  .723      &   .292   &  1758   &  .136  &  .717  &  .230  \\
\cdashline{2-18}
&  i-biLSTM     &  404  &     {.487}     &   .582    &    .531    &  940   &   .245      &  .682      &  .360   &  1205 &   .185      &  .661      &  .289     &  2146 &  .128  &  .816  &  .222  \\
&  i-biRNN     &  288  &     .566     &   .482    &    .521    &  462  &   .376      &  .515      &  .435  &  556  &   .318      &  .524      &  .396    &  601  &  ..296  &  .525  &  .378 \\
& iDepNN-SDP    &  335  &    .537     &   .531    &    .534   &  633   &   .319     &  .598      &  .416  &  832  &   .258      &  .634      &  .367   &  941   &  .228  &  .633  &  .335 \\ 
& iDepNN-ADP   &  309  &      \underline{.538}     &   .493    &    .514    &  485  &    \underline{.365}      &  .525      &  .431  &  572   &    \underline{.320}      &  .542      &  .402    &  603  &   \underline{.302}  &  .540  &  .387  \\ 
\hline 
\multirow{5}{*}{$k \le 3$}   &  SVM      &  507   &    .458     &  {\bf  .686}    &    {.549}    &  1172  &   .234      &     {\bf  .811}  &  .363  &  1629  &    .186     &  {\bf .894}      &   .308   &  1874  &  .162  &  {\bf .897}  &  .275 \\ 
  &  graphLSTM       &  429   &    .491     &   .624    &    .550    &  1082   &   .230      &      .740  &  .351  &  1673   &    .167     
&  .833      &   .280   &  2126   &  .124  &  .787  &  .214  \\
\cdashline{2-18}
&  i-biLSTM      &  417   &    .478     &   .582    &    .526    &  1142   &   .224      &      .758  &  .345  &  1218   &    .162     
&  .833      &   .273   &  2091   &  .128  &  .800  &  .223  \\
&  i-biRNN     &  405  &     {.464}     &   .559    &    .507    &  622  &   {.324}      &  .601      &  .422  &  654  &   {.310}      &  .604      &  .410   &  655   &  {.311}  &  .607  &  .410   \\ 
& iDepNN-SDP    &  351  &    .533     &   .552    &    .542    &  651  &   .315      &  .605      &  .414  &  842  &   .251      &  .622      &  .357   &  928   &  .227  &  .622  &  .333  \\ 
& iDepNN-ADP   &  313  &     \underline{.553}     &   .512    &    .532    &  541  &   \underline{.355}      &  .568      &  .437  &  654  &   \underline{.315}      &  .601      &  .415   &  687   &  \underline{.300}  &  .601  &  .401 \\
\hline \hline
\multirow{1}{*}{$\bf {k \le 1}$}     &  {\it ensemble}     &  480  &     {.478}     &   .680    &    {\bf .561}    &  837  &   .311      &  .769      &  .443  &  1003  &   .268      &  .794      &  .401   &  1074   &  .252  &  .797  &  .382  
\end{tabular}}
\caption{BioNLP ST 2016  Dataset: Performance of the intra-and-inter-sentential training/evaluation for different $k$. 
\underline{Underline}: Better precision by {\it iDepNN-ADP} over {\it iDepNN-SDP}, graphLSTM and SVM. {\bf Bold}: Best in column. ${\bf pr}$: Count of predictions} 
\label{BioNLPstateoftheart2016}
\end{table*}

\textbf{Experimental Setup.}
For MUC6, we use the pretrained GloVe \cite{pennington2014glove} embeddings (200-dimension). For the BioNLP datasets, we 
use 200-dimensional embedding\footnote{\url{http://bio.nlplab.org/}} vectors from  six billion words of  biomedical text \cite{moen2013distributional}. We randomly initialize a 5-dimensional vectors for PI and POS. 
We initialize the recurrent weight matrix to identity  and biases to zero. 
We use the macro-averaged $F_1$ score (the official evaluation script by SemEval-2010 Task 8 \cite{hendrickx2010semeval}) on the development set 
to choose hyperparameters (see supplementary).  
To report results on BioNLP ST 2016 test set, 
we use the official web service\footnote{\url{http://bibliome.jouy.inra.fr/demo/BioNLP-ST-2016-Evaluation/index.html}}.


\textbf{Baselines.}
\citeauthor{swampillai2010inter}'s  \shortcite{swampillai2010inter}
annotation of MUC6 intra- and inter-sentential relationships
is not available.
They investigated SVM with dependency and linguistic features for relationships spanning sentence boundaries. 
In BioNLP shared tasks, the top performing systems are SVM-based and limited to relationships within single sentences.   
As an NN baseline, we also develop Connectionist biRNN \cite{vu2016combining} that spans sentence boundaries; 
we refer to it as i-biRNN (architecture in  supplementary). Similarly, we also investigate using a bidirectional LSTM (i-biLSTM). 
As a competitive baseline in the inter-sentential relationship extraction, we run\footnote{hyperparameters in supplementary} graphLSTM \cite{peng2017cross}.   
This work compares SVM and graphLSTM with i-biRNN, i-biLSTM, iDepNN-SDP and iDepNN-ADP for different values of the sentence range parameter $k$  
(the distance in terms of the number of sentences between the entity pairs for a relation) , i.e., 
$k$ ($=0$, $\le 1$, $\le 2$ and $\le 3$).  

\textbf{Contribution of different components.}
Table \ref{SVMiDepNNFeatureAnalysis} shows the contribution of each feature, where both training and evaluation 
is performed over relationships within and across sentence boundaries for sentence range parameter $k$$\le$$1$. 
Note: iSDP+Dependency refers to iDepNN-ADP structure 
that exhibits  a better precision, $F_1$ and balance in precision and recall, compared to SVM. 
See supplementary  for feature analysis on BioNLP ST 2011 / 2013.

\subsection{State-of-the-Art Comparisons}
{\bf BioNLP ST 2016 dataset:}
Table \ref{BioNLPstateoftheart2016} shows the performance
of \{SVM, graphLSTM\} vs \{i-biRNN, iDepNN-SDP, iDepNN-ADP\}
for relationships within and across sentence boundaries.
{\it Moving left to right} for {\it each} training parameter, the recall increases while precision decreases due to  
increasing noise with larger $k$.   
In the inter-sentential evaluations ($k \le 1, \le 2, \le3$ columns) for {\it all} 
the training parameters, the iDepNN variants outperform both SVM and graphLSTM in terms of $F_1$ and  maintain a better precision as well as balance 
in precision and recall with increasing $k$; e.g., 
 at $k \le 1$ (train/eval),  precision and $F_1$ are  
($.402$ vs $.226$) and ($.449$ vs $.336$), respectively for (iDepNN-ADP vs graphLSTM).   
We find that SVM mostly leads in recall.  

\begin{table}[t]
\centering
\def\arraystretch{1.2}
\resizebox{0.47\textwidth}{!}{%
\begin{tabular}{c|cccc|cccc}
\multirow{3}{*}{{\it threshold}} &  \multicolumn{8}{c} {{\it ensemble} (train $k \le 1$ and evaluation $k=0$)}\\ \cline{2-9}
&  \multicolumn{4}{c|}{\textit{Dev  (official scores)}} & \multicolumn{4}{c}{\textit{Test} (official scores)} \\
&  {\bf pr} & $P$ & $R$ & $F_1$ & {\bf pr} & $P$ & $R$ & $F_1$ \\ \hline
$p \ge 0.85$ &  160 & .694 &  .514 &  .591   & 419 &  .530 &  .657 &  {\bf .587} \\
$p \ge 0.90$ &  151 & .709 &  .496 &  .583   & 395 &  .539 &  .630 &  {.581} \\
$p \ge 0.95$ &  123 & .740 &  .419 &  .535   & 293 &  .573 &  .497 &  .533 
\end{tabular}}
\caption{Ensemble scores at various thresholds 
for BioNLP ST 2016 dataset. $p$: output probability}
\label{thresholdanalysis}
\end{table}

\def\smaller{0.075cm}

\begin{figure*}[t]
\begin{tabular}{ll}
\begin{tabular}{c}
   \includegraphics[scale=.41] {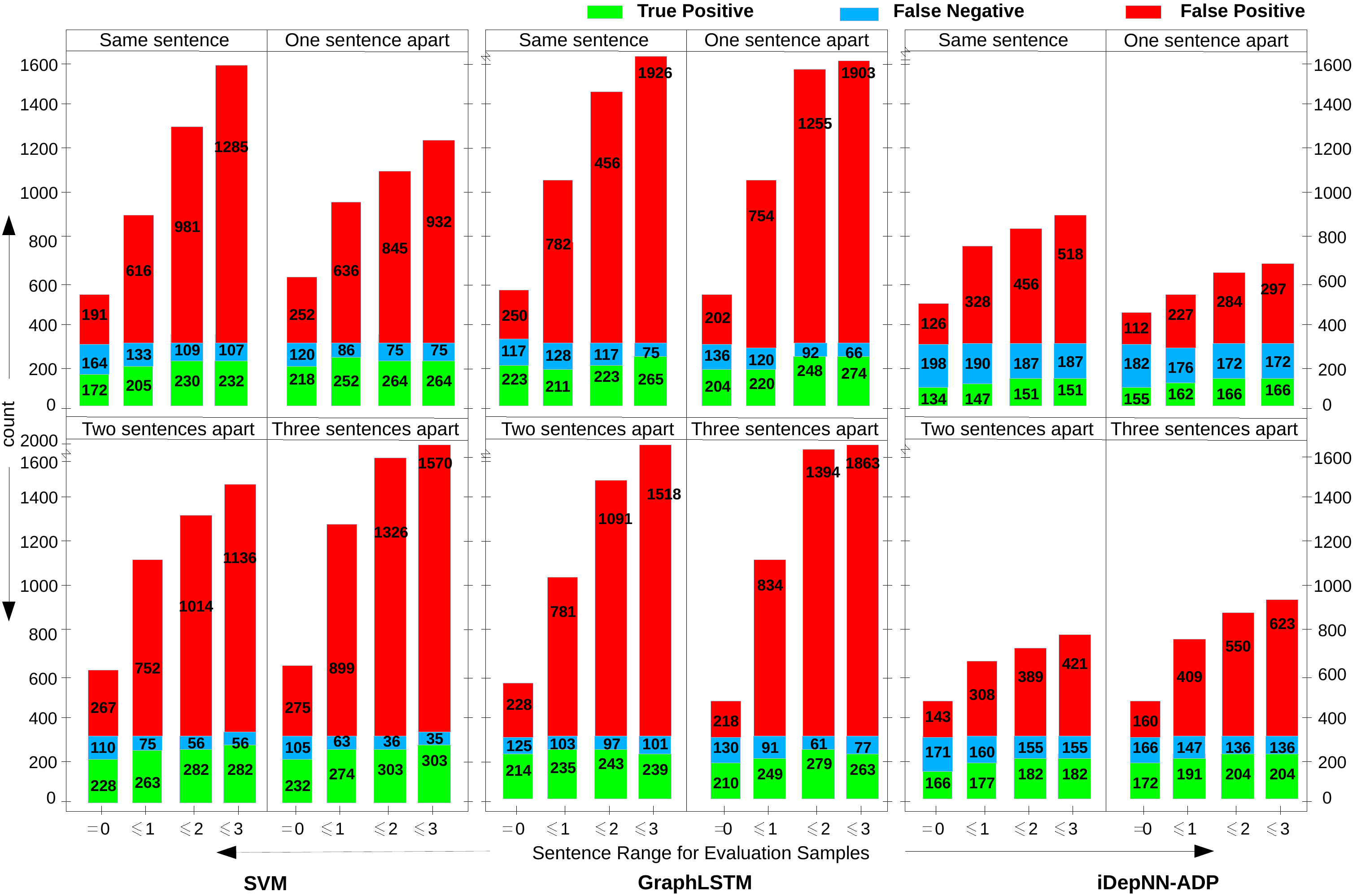} 
\end{tabular}
&
{\scriptsize
\begin{tabular}{l|c@{\hspace{\smaller}}c@{\hspace{\smaller}}c@{\hspace{\smaller}}c}
\multirow{2}{*}{\textbf{Teams}}          & \multicolumn{4}{c}{\textbf{Lives\_In}} \\ 
& $F_1$ &R & P & pr\\ \hline
{\it this work}  & {\bf .587}         & {\bf .657}        & .530         & 419\\
VERSE    & .558         & .615        & .510         & 408 \\ 
TurkuNLP    & .522     & .448        & .626        & 243 \\ 
LIMSI    & .485         & .646        & .388        & 559 \\ 
HK       & .474         & .392        & .599        & 222\\ 
WhuNlpRE & .471         & .407        & .559        & 247 \\
UMS       & .463 & .399 &  .551 &  245 \\
DUTIR	   & .456	 & .382	 & .566	 & 228 \\      
WXU	 & .455	 & .383	 & .560	 & 232 
\end{tabular}}
\end{tabular}
    \caption{Left: SVM, graphLSTM \& iDepNN-ADP on BioNLP ST 2016: Performance analysis on relations that span sentence boundaries, with different sentence range parameters
        Right:
BioNLP 2016 ST dataset (official results on test set): Comparison with the published systems in the BioNLP ST, where $pr$ is the count of predictions.   
This work demonstrates a better balance in precision and recall, and  achieves the highest $F_1$ and {\it recall}. 
We extract $419$ predictions within and across sentence boundaries, 
which is closer to the count of gold predictions, i.e., $340$ \cite{deleger2016overview}.
\label{systemperformanceinsharedtask}}
\end{figure*}


\begin{table*}[t]
\centering
\small
\def\arraystretch{1.2}
\resizebox{0.95\textwidth}{!}{%
\setlength\tabcolsep{2.5pt}
\begin{tabular}{c||ccc|ccc|ccc|ccc||ccc|ccc|ccc|ccc}
\multirow{3}{*}{\bf Model}  & \multicolumn{12}{c||}{\it Dataset: BioNLP ST 2013}     & \multicolumn{12}{c}{\it Dataset: BioNLP ST 2011}\\ \cline{2-25}    
 & \multicolumn{3}{c|}{\bf $k=0$} & \multicolumn{3}{c|}{\bf $k \le 1$} & \multicolumn{3}{c|}{\bf $ k \le 2$} & \multicolumn{3}{c||}{$k \le 3$} & \multicolumn{3}{c|}{\bf $k=0$} & \multicolumn{3}{c|}{\bf $k \le 1$} & \multicolumn{3}{c|}{\bf $ k \le 2$} & \multicolumn{3}{c}{$k \le 3$}\\ 
                    & $P$      & $R$     & $F_1$     & $P$       & $R$      & $F_1$     & $P$       & $R$      & $F_1$    & $P$       & $R$      & $F_1$    & $P$      & $R$     & $F_1$     & $P$       & $R$      & $F_1$     & $P$       & $R$      & $F_1$    & $P$       & $R$      & $F_1$  \\ \hline 
SVM                     & .95	& .90	& .92	& .87	& .83	& .85	& .95	& .90	& .92	& .95	& .90	& .92                                   & { .98}	& .96	& .97	& .87	& .87	& .87	& .95	& .94	& .94	& .91	& .88	& .90       \\
graphLSTM          & {.98}	       & .97	& .97	& .94	& {\bf .95}	& .94	& .95& .89	& .92	& .90	& {\bf .97}	& {\bf .93}     & {\bf .99}	& .99	& {\bf .99}& .95	& {\bf .98} & {\bf .96}	& .95	& .97& .96& .96& .92 	& .93             \\ \cdashline{1-25}
i-biLSTM               & {.98}	       & .97	& .97	& .96	& {\bf .95}	& {\bf .95}	& .93	& {\bf .96}	& .94	& .91& .93	& .92       & {\bf .99}	& .99	& {\bf .99	}& .95& {\bf .98}	& {\bf .96} 	& .96& .97	& .96& .95	& .92 	& .93           \\
i-biRNN              & .95 & .94	& .94	& .93	& .90	& .91	& .94	& .92	& .93	& .94	& .84	& .89                                       & .97	& .99	& .98		& .88 & .94	& .90	& .92	& .94	& .93	& {\bf .96}	& {\bf .96}	& {\bf .96} 	\\ 
iDepNN-SDP       & .94	& .96	& .95	& .94	& {\bf .95} & .94	& .87	& .92	& .89	& .91	& .94	& .92                  & .97	& {\bf .99}	& .98	& {\bf .96}	& .92	& .93	& {\bf .97}	& .97	& {\bf .97}	& .94	& .91	& .92  \\ 
iDepNN-ADP       & \underline{\bf .99}	& {\bf .98}	& {\bf .99}	& \underline{\bf .97}	& .94	& {\bf .95}	& \underline{\bf .98}	& .95 	& {\bf .96}	&\underline{\bf .96}	& .91	& {\bf .93}             & .97	& .97	& .97	& .93	& .96	& 94	& .92	& {\bf .98}	& .95  & .93	& .94	& .93 
\end{tabular}}
\caption{BioNLP ST 2011 and 2013 datasets: Performance for training ($k \le 1$) and evaluation for different $k$. 
\underline{Underline}: Better precision in iDepNN-ADP than iDepNN-SDP, graphLSTM, i-biLSTM, i-biRNN and SVM. {\bf Bold}: best in column.} 
\label{BioNLP20112013stateoftheart}
\end{table*}

\begin{table}[t]
\centering
\def\arraystretch{1.2}
\resizebox{.49\textwidth}{!}{
\setlength\tabcolsep{2.5pt}
\begin{tabular}{c|c||ccc|ccc|ccc}
\multirow{1}{*}{\bf train}   & \multirow{3}{*}{\bf Model}  &  \multicolumn{9}{c}{\bf Evaluation for different $k$}\\ \cline{3-11}
 \multirow{1}{*}{\bf param}     &   & \multicolumn{3}{c|}{\bf $k=0$} & \multicolumn{3}{c|}{\bf $k \le 1$}  & \multicolumn{3}{c}{\bf $k \le 3$}\\ 
        &   & $P$      & $R$     & $F_1$      & $P$       &  $R$      & $F_1$  & $P$       &  $R$      & $F_1$ \\ \hline
\multirow{4}{*}{$k=0$}   
&  SVM                                                    & .796	& .765	& .760	& .775	& .762	    & .759	& .791	& .779	& .776  \\ 
&  graphLSTM                                          & .910	& .857	& .880	& .867	& .897	     & .870	& .870	& .867	& .870  \\ \cdashline{2-11}
&  i-biLSTM                                             & .917	& .837	& .873	& .833	& .896	& .863	& .853	& .87	0 & .863\\ 
& i-biRNN                                                & .875	& .859	& .864	& .828	& .822	& .824	 & .830	& .827	& .827\\
& iDepNN-SDP                                       & .958	& .948	& .952	& .934	& .928	& .930	& .935	& .930	& .932\\
& iDepNN-ADP                                       & .933	& .927	& .929	& .924	& .920	& .921	& .930	& .927	& .927
\\ \hline
\multirow{4}{*}{$k \le 1$}   
&  SVM                                                   & .815	& .772	& .769	 & .808	& .768	& .764 	      & .802	& .775	& .770\\ 
&  graphLSTM                                          & .730	& .900	& .783	& .727	& .907	& .773	& .730	& .913	& .770 \\ 
\cdashline{2-11}
&  i-biLSTM                                            & .760	& .880	& .780	& .670	& .950	& .767	      & .697	& .937	& .770\\ 
& i-biRNN                                                & .925	& .934	& .927  & .870	& .872	& .860	                & .868	& .866	& .858\\
& iDepNN-SDP               & .949	& .945	& .946	& .928	& .926  & .926                                         & .934	& .932	& .932\\
& iDepNN-ADP                                        &  \underline{.961}	& .955	& .957	&  \underline{\bf .937}	& {\bf .934}	& {\bf .935}   & \underline{\bf .942}	& {\bf .940}	& {\bf .940} \\ \hline
\multirow{4}{*}{$k \le 3$}   
&  SVM                                                   & .840	& .785	& .779	& .816	& .779	& .774	       & .822	& .788	& .781\\ 
&  graphLSTM                                          & .737	& .907	& .783	& .703	& .927	& .773	& .710	& .927	& .767\\ 
\cdashline{2-11}
&  i-biLSTM                                          & .720	& .920	& .780	& .680	& .943	& .770	& .700	& .932	& .770\\  
& i-biRNN                                                & .944	& .934	& .938	& .902  & .890	& .895	        & .926	& .923	& .924\\
& iDepNN-SDP                                       & .956	& .947	& .951	& .920	& .916	& .917	& .939	& .938	& .936 \\
& iDepNN-ADP                                       & \underline{\bf .965}	& {\bf .963}	& {\bf .963}	& \underline{.933}	& .933	& .931   & \underline{.939}	& .938	& .936
\end{tabular}}
\caption{MUC6 Dataset: Performance  over the intra- and inter-sentential training and evaluation for different $k$. 
\underline{Underline} signifies better precision by iDepNN-ADP over iDepNN-SDP, graphLSTM, i-biLSTM, i-biRNN and SVM. {\bf Bold} indicates the best score column-wise.} 
\label{Muc6stateoftheart2016}
\end{table}

In comparison to graphLSTM, i-biRNN and i-biLSTM, we observe that iDepNN-ADP offers precise structure in relation extraction within and across sentence boundaries. 
For instance, at training $k \le 1$ and evaluation $k=0$, iDepNN-ADP reports precision of  $.570$  vs  $.489$ and $.561$ in i-biRNN and iDepNN-SDP, respectively. 
During training at $k \le 1$, iDepNN-SDP and iDepNN-ADP report better $F_1$ than i-biRNN for evaluations at all $k$, 
suggesting that the shortest and augmented paths provide useful dependency features via recurrent and recursive compositions, respectively.   
Between the proposed architectures iDepNN-SDP and iDepNN-ADP, the former achieves higher recall for all $k$.  
We find that the training at $k \le 1$ is optimal for intra- and inter-sentential relations over development set (see  supplementary).  
We also observe that i-biRNN establishes a strong NN baseline for 
relationships spanning sentences. The proposed architectures consistently outperform graphLSTM in both precision and $F_1$ across sentence boundaries. 

\textit{Ensemble}: We exploit the precision and recall bias of the different models via an ensemble approach, similar to TurkuNLP \cite{mehryary2016deep} 
and UMS \cite{deleger2016overview} systems that combined predictions from SVM and NNs.   
We aggregate the prediction outputs of the neural (i-biRNN, iDepNN-SDP and iDepNN-ADP) and non-neural (SVM) classifiers, i.e.,  
a relation to hold if any classifier has predicted it. We perform the {\it ensemble} scheme on the development and official test sets 
for intra- and inter-sentential (optimal at $k \le 1$) relations. 
Table \ref{BioNLPstateoftheart2016} shows the ensemble scores on the official test set for relations  within and across 
sentence boundaries, where {\it ensemble} achieves the highest $F_1$ ($.561$) over individual models.

\textit{Confident Extractions}: We consider the high confidence prediction outputs by the different models participating in the {\it ensemble}, 
since it lacks precision ($.478$). Following \citeauthor{peng2017cross} \shortcite{peng2017cross}, 
we examine three values of the output probability 
$p$, i.e., ($\ge 0.85, 0.90$ and $0.95$) of each model in the {\it ensemble}. 
Table \ref{thresholdanalysis} shows the {\it ensemble} performance on the development and official test sets, where the predictions with $p \ge 0.85$ 
achieve the state-of-the-art performance and rank us at the
top out of 11 systems
(Figure \ref{systemperformanceinsharedtask}, right).

\textit{This Work vs Competing Systems in BioNLP ST 2016}:
As shown in Figure  \ref{systemperformanceinsharedtask} (right), we rank at the top  and achieve 
a gain of 5.2\% ($.587$ vs $.558$) in $F_1$ compared to VERSE.  
We also show a better balance in precision and recall, and report the highest recall compared to all other systems.  
Most systems do not attempt to predict relations spanning sentences. 
The most popular algorithms are SVM (VERSE, HK, UTS, LIMSI) and NNs (TurkuNLP, WhuNlpRE, DUTIR). 
UMS combined predictions from an SVM and an NN. 
Most systems rely on syntactic parsing, POS,  
word embeddings and entity recognition features 
(VERSE, TurkuNLP, UMS, HK, DUTIR, UTS). 
VERSE and TurkuNLP obtained top scores on intra-sentential relations relying on the dependency path features between entities; 
however they are limited to intra-sentential relations. TurkuNLP employed an ensemble of 15 different LSTM based classifiers.  
DUTIR is based on CNN for intra-sentential relationsips.  
LIMSI is the only system that considers inter-sentential relationships during training; however it is SVM-based and used additional manually annotated training data,  
linguistic features using biomedical resources (PubMed, Cocoa web API, OntoBiotope ontology, etc.) and post-processing to annotate biomedical abbreviations. 
We report a noticeable gain of 21\% ($.587$ vs $.485$) in $F_1$ with an improved precision and recall over LIMSI. 

{\bf BioNLP ST 2011 and 2013 datasets:} Following the BioNLP ST 2016 evaluation, 
we also examine two additional datasets from the same domain.   
iDepNN-ADP (Table \ref{BioNLP20112013stateoftheart}) is the leading performer in terms of precision and $F_1$ within and across boundaries for BioNLP ST 2013. 
Examining BioNLP ST 2011, the iDepNN variants lead both SVM and i-biRNN for $k \le 1$ and   $k \le 2$. 

{\bf MUC6 dataset:} Similar to BioNLP ST 2016, we perform training and evaluation of 
SVM, i-biRNN, iDepNN-SDP and iDepNN-ADP for different sentence range with best feature combination (Table \ref{SVMiDepNNFeatureAnalysis})
using MUC6 dataset.   
Table \ref{Muc6stateoftheart2016} shows that both iDepNN variants consistently outperform graphLSTM and SVM for relationships  
within and across sentences. For within ($k$$=$$0$) sentence evaluation, iDepNN-ADP reports $.963$ $F_1$, compared to $.779$ and $.783$ by SVM and graphLSTM, respectively. 
iDepNN-ADP is observed more precise than iDepNN-SDP and graphLSTM with increasing $k$, e.g., at $k$$\le$$3$. 
Training at sentence range $k$$\le$$1$  is found optimal in extracting inter-sentential relationships. 

\subsection{Error Analysis and Discussion}\label{erroranalysis}
In Figure \ref{systemperformanceinsharedtask} (left), we analyze predictions using different values of sentence range $k$ (=0, $\le$1, $\le$2 and $\le$3) during both training and evaluation of SVM, graphLSTM and iDepNN-ADP for BioNLP ST 2016 dataset.  
For instance, an SVM (top-left) is trained for intra-sentential ({\it same sentence}) relations, while iDepNN-ADP (bottom-right) for both intra- and inter-sentential spanning three sentences ({\it three sentences apart}).  
We show how the count of true positives (TP), false negatives (FN) and false positives (FP) varies with $k$. 

Observe that as the distance of the relation increases, the classifiers predict larger ratios of false positives to true positives. 
However, as the sentence range increases, iDepNN-ADP outperforms both SVM and graphLSTM due to fewer false positives (red colored bars).  
On top, the ratio of FP to TP is better in iDepNN-ADP than graphLSTM and SVM for all values of $k$. 
Correspondingly in Table \ref{BioNLPstateoftheart2016}, iDepNN-ADP reports better  precision and balance between precision and recall, signifying its robustness to noise
 in handling inter-sentential relationships.  

{\bf iDepNN vs graphLSTM}: 
\citeauthor{peng2017cross} \shortcite{peng2017cross} focuses on general relation extraction framework using graphLSTM with challenges 
such as potential cycles in the document graph leading to
expensive model training and difficulties in convergence due
to loopy gradient backpropagation.  Therefore, they further
investigated different strategies to backpropagate
gradients. 
The graphLSTM introduces a number of parameters with a number of
edge types and thus, requires abundant supervision/training
data.  On other hand, our work introduces simple and robust
neural architectures (iDepNN-SDP and iDepNN-ADP), where the
iDepNN-ADP is a special case of document graph in form of a
parse tree spanning sentence boundaries. We offer a smooth
gradient backpropagation in the complete structure (e.g., 
in iDepNN-ADP via recurrent and recursive hidden
vectors) that is more efficient than graphLSTM due to non-cyclic
(i.e., tree) architecture.  We have also shown that
iDepNN-ADP is robust to false positives and maintains a
better balance in precision and recall than graphLSTM for
inter-sentential relationships
(Figure \ref{systemperformanceinsharedtask}).

\section{Conclusion}
\label{conclusion}
We have proposed to classify relations 
between entities within and across sentence boundaries 
by modeling the inter-sentential shortest and augmented dependency
paths within a novel neural network, 
named as inter-sentential Dependency-based Neural Network (iDepNN) that 
takes advantage of both recurrent and recursive neural networks  
to model the structures in the intra- and inter-sentential relationships. 
Experimental results on four datasets from  newswire and medical domains 
have demonstrated that iDepNN is robust to false positives, 
shows 
a better balance in precision and recall and 
achieves the state-of-the-art performance in extracting relationships within and across sentence boundaries. 
We also perform better than 11 teams participating  in the BioNLP shared task 2016. 



\section*{Acknowledgments}
We thank anonymous reviewers for their review comments. 
This research was supported by Bundeswirtschaftsministerium ({\tt bmwi.de}), grant 01MD15010A (Smart Data Web) 
at Siemens AG- CT Machine Intelligence, Munich Germany.

\small

\bibliography{aaai19}
\bibliographystyle{aaai19}



\section*{Supplementary material}

# Supplementary Material

## AAAI Press

Association for the Advancement of Artificial Intelligence
2275 East Bayshore Road, Suite 160
Palo Alto, California 94303

## Architecture of i-biRNN

Figure 1 illustrates the architecture of inter-sentential Connectionist Bi-directional Recurrent Neural Network (i-biRNN). Consider two sentences S1 and S2, where S1 consists of words $[v_1^1, [v_2^1]_{e1}, v_3^1, v_4^1]$ and S2 consists of words $[w_1^2, w_2^2, [w_3^2]_{e2}, w_4^2]$. $v_2^1$ and $w_3^2$ are *entity1* and *entity2* respectively spanning sentence boundary.

The input to the i-biRNN is the concatenation of S1 and S2 where;

- $\mathbf{V}$ : the weights matrix between hidden units and input in forward and backward network used to condition the input word vector, $\mathbf{x_t}$

- $\mathbf{W_f}$ : the weights matrix connecting hidden units in forward network

- $\mathbf{W_b}$ : the weights matrix connecting hidden units in backward network

- $\mathbf{h_f}$ : the forward hidden unit computed at time step, t, accumulating the semantic meaning given the input word $\mathbf{x_t}$ and history

- $\mathbf{h_b}$ : the backward hidden unit computed at time step, t, accumulating the semantic meaning given the input word $\mathbf{x_t}$ and the future context, $\mathbf{h_{b_{t+1}}}$, that is accumulated and conditioned on the future words.

- $\mathbf{W}$ : the recurrent weight matrix connecting combined hidden units, $\mathbf{h_t}$ through time;

- $\mathbf{h}_t$ : the hidden vector that accumulates the semantics obtained from the combination or sum of forward and backward units at time step, t

- N : total number of words in S1 and S2

- $\mathbf{U}$: the output weight matrix, connecting to the softmax layer

## Data Description

The BioNLP shared task (2011 and 2013) consists of extracting bacteria localization events of given species i.e., *Localization* or *PartOf* relations. For *Location*, the participating entity types are Bacterium and type Localization (Host

| Parameter | Value(s) |
|---|---|
| Training epoch | 500 |
| Hidden size | 100, 200 |
| initial learning rate | 0.001, 0.005, 0.010 |
| activation | tanh, cappedrelu |
| context window size | 1, 3 |
| L2 weight | 0.0001 |
| mini batch size | 1 |
| optimizer | gradient descent |
| gradient norm clipstyle | rescale |
| gradient norm cutoff | 5.0, 10.0, 20.0 |
| word embedding dimension | 200 |
| PI/POS embedding dimension | 5 |
| sentence range, $k$ | $=0, \leqslant 1, \leqslant 2, \leqslant 3$ |
| weight initialization | identity |
| bias initialization | zero |
| $\mathbf{W}$ sharing in bi-RNN | True |
| $\mathbf{W}$ sharing in ADP and SDP | True, False |

Table 1: Hyperparameters for BRNN and iDepNN

| HostPart | Geographic | Environment | Food | Soil | Water), while *PartOf* has *HostPart* and *Host* entity types. The BioNLP ST 2016 focuses on extraction of *Lives_in* events among *Bacteria*, *Habitat* and *Geographical* entities.

## Hyperparameters

See Table 1. We run graphLSTM for 100 iterations with learning rate of 0.02, batch size 10 and 150 hidden dimension.

## Feature Analysis on BioNLP ST 2011 and 2013 datasets

See Table 2.

## Development Scores (official) on BioNLP ST 2016 data set to determine the value of training parameter, $k$

See Table 3.

| Dataset: BioNLP ST 2013 | | | | | | |
|---|---|---|---|---|---|---|
| **Features** | **SVM** | | | **iDepNN** | | |
| | P | R | F1 | P | R | F1 |
| iSDP | .848 | .740 | .769 | .911 | .875 | .889 |
| + PI + ET | .853 | .786 | .808 | .915 | .921 | .918 |
| + POS | .934 | .904 | .917 | .938 | .945 | .942 |
| + Dependency | .937 | .926 | .931 | .970 | .938 | **.951** |
| Dataset: BioNLP ST 2011 | | | | | | |
| **Features** | **SVM** | | | **iDepNN** | | |
| | P | R | F1 | P | R | F1 |
| iSDP | .830 | .777 | .794 | .919 | .933 | .926 |
| + PI + ET | .831 | .790 | .803 | .931 | .916 | .923 |
| + POS | .915 | .918 | .915 | .955 | .922 | .938 |
| + Dependency | .930 | .931 | .930 | .928 | .957 | **.942** |

Table 2: SVM Vs iDepNN: Performance of different features used in inter-sentential ($k \leqslant 1$) training and inter-sentential ($k \leqslant 1$) evaluation. *iSDP + dependency* refers to *iDepNN-ADP*.

| train param | Model | **Development scores at different values of $k$** | | | | | | | |
|---|---|---|---|---|---|---|---|---|---|
| | | $k = 0$ | | | | $k \leqslant 1$ | | | |
| | | **pr** | **F1** | **R** | **P** | **pr** | **F1** | **R** | **P** |
| $k \leqslant 1$ | i-biRNN | 113 | **.497** | .378 | .726 | 132 | .591 | .475 | .781 |
| | iDepNN-SDP | 139 | .524 | .431 | **.669** | 167 | **.641** | **.568** | .737 |
| | iDepNN-ADP | 138 | **.536** | .438 | **.688** | 161 | **.630** | .549 | .740 |
| $k \leqslant 2$ | i-biRNN | 129 | .493 | .393 | .659 | 145 | .613 | .511 | .764 |
| | iDepNN-SDP | 149 | .531 | **.449** | .651 | 180 | .598 | .545 | .661 |
| | iDepNN-ADP | 142 | .534 | .443 | .675 | 161 | .619 | .539 | .726 |
| $k \leqslant 3$ | i-biRNN | 112 | .514 | .388 | .759 | 129 | .603 | .482 | **.806** |
| | iDepNN-SDP | 142 | .506 | .417 | .641 | 173 | .610 | .548 | .688 |
| | iDepNN-ADP | 139 | .530 | .435 | .676 | 158 | .626 | .539 | .747 |

Table 3: BioNLP ST 2016 development set: Performance comparison to determine the value of training parameter $k$. Observe that the parameter $k \leqslant 1$ is optimal during training and evaluation (on development set) for relationships within and across sentence boundaries. $P$, $R$ and $F1$ are official scores.

Zhang, M.; Zhang, J.; Su, J.; and Zhou, G. 2006. A composite kernel to extract relations between entities with both flat and structured features. In *Proceedings of the 21st International Conference on Computational Linguistics and the 44th annual meeting of the Association for Computational Linguistics*, 825–832. Association for Computational Linguistics.

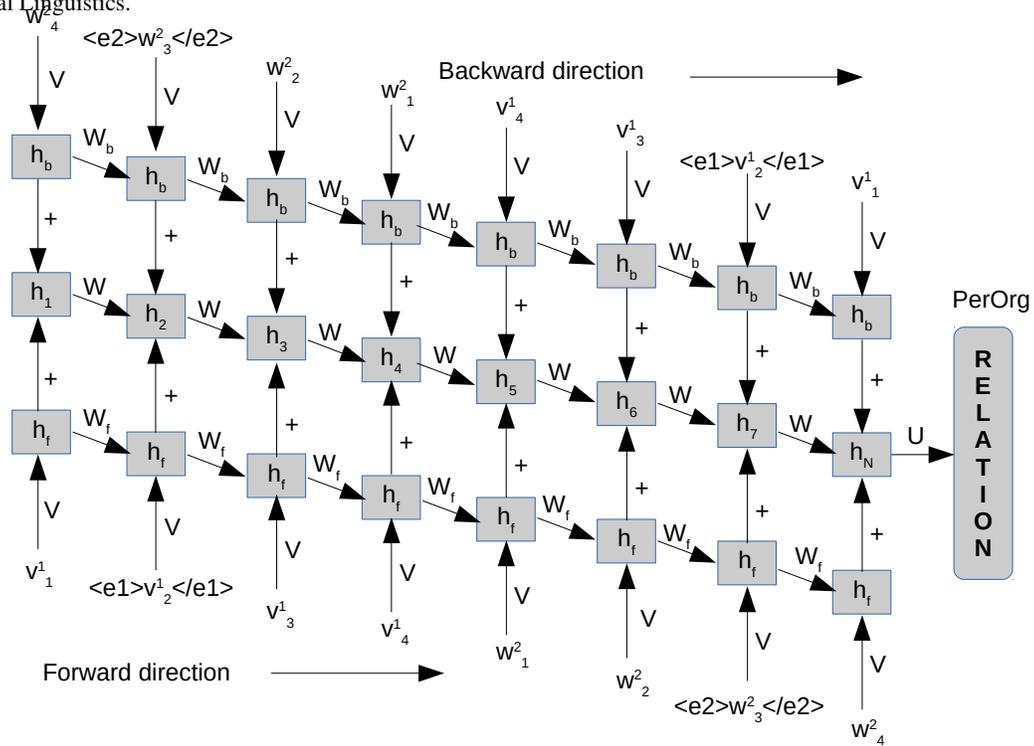

Figure 1: Architecture of i-biRNN. In our structures, we share the weights in forward, backward and combined networks, in order to reduce parameters, i.e., **W** is being shared in the three networks.



\end{document}